\documentclass{article}
\usepackage{graphicx} 
\usepackage{subcaption}

\usepackage{wrapfig}
\usepackage{amsfonts}
\usepackage{amsmath}
\usepackage{array}

\usepackage{bbm}
\usepackage[final]{corl_2024} 

\title{Dynamic Obstacle Avoidance with Bounded Rationality Adversarial Reinforcement Learning}

\author{
  Jose-Luis Holgado-Alvarez$^1$, Aryaman Reddi$^{2,3}$,  Carlo D'Eramo$^{1,2,3}$\\
   Center for Artificial Intelligence and Data Science, University of W\"{u}rzburg, Germany$^1$\\
   Technische Universität Darmstadt, Germany$^2$ \\
   Hessian.ai, Germany$^3$\\
    \texttt{jose-luis.holgado-alvarez@uni-wuerzburg.de} \\
}

\begin{document}
\maketitle


\begin{abstract}
        Reinforcement Learning (RL) has proven largely effective in obtaining stable locomotion gaits for legged robots. However, designing control algorithms which can robustly navigate unseen environments with obstacles remains an ongoing problem within quadruped locomotion. To tackle this, it is convenient to solve navigation tasks by means of a hierarchical approach with a low-level locomotion policy and a high-level navigation policy. Crucially, the high-level policy needs to be robust to dynamic obstacles along the path of the agent. In this work, we propose a novel way to endow navigation policies with robustness by a training process that models obstacles as adversarial agents, following the adversarial RL paradigm. Importantly, to improve the reliability of the training process, we bound the rationality of the adversarial agent resorting to quantal response equilibria, and place a curriculum over its rationality. We called this method Hierarchical policies via Quantal response Adversarial Reinforcement Learning (\textbf{Hi-QARL}). We demonstrate the robustness of our method by benchmarking it in unseen randomized mazes with multiple obstacles. To prove its applicability in real scenarios, our method is applied on a Unitree GO1 robot in simulation.
\end{abstract}
    
\keywords{locomotion, skills decomposition, robustness, bounded rationality} 


\begin{figure}[h!]
    \centering
    \begin{subfigure}{0.32\textwidth}
        \centering
        \includegraphics[width=\linewidth]{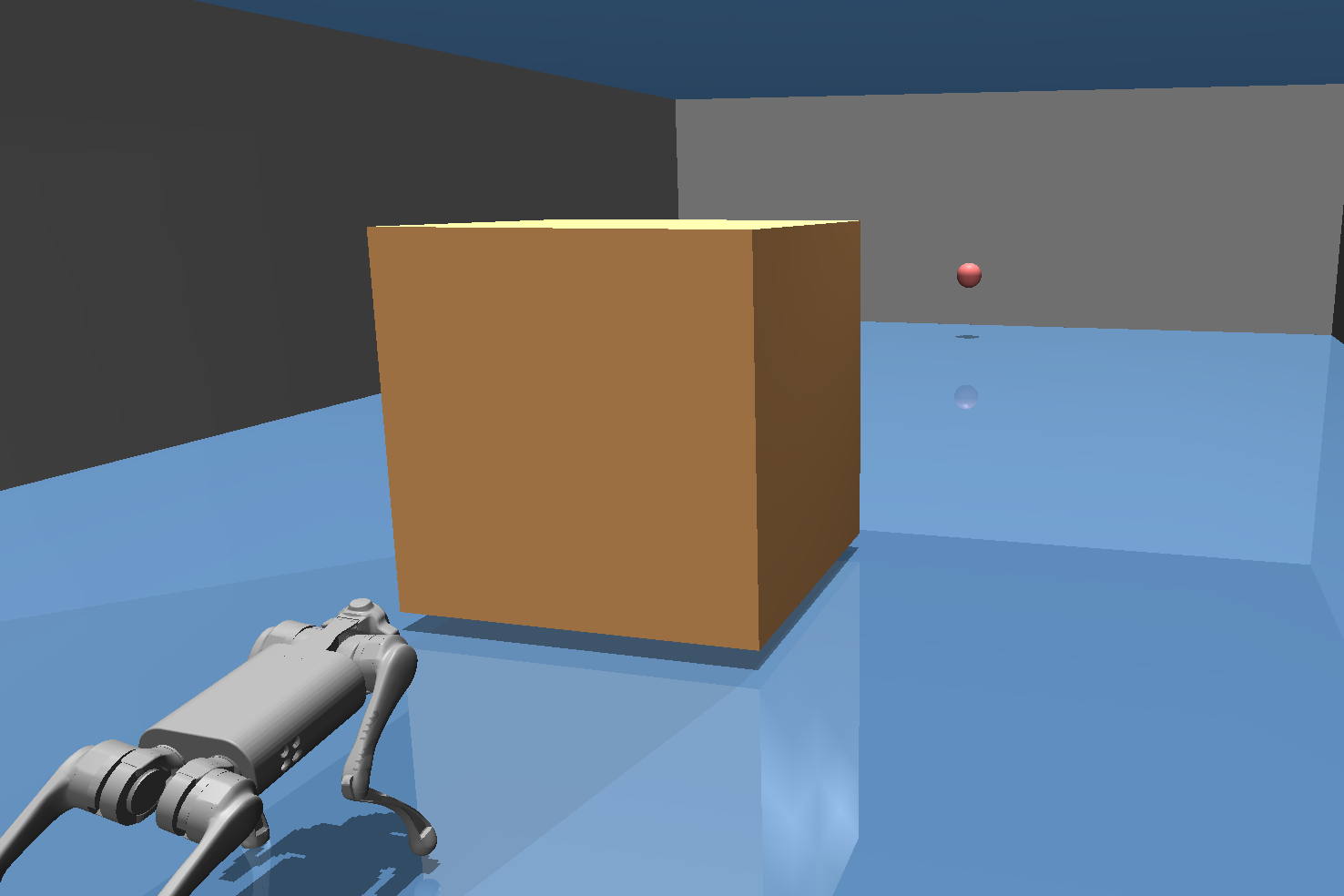}
        \caption{Approaching the obstacle.}
        \label{fig:image1}
    \end{subfigure}
    \hfill
    \begin{subfigure}{0.32\textwidth}
        \centering
        \includegraphics[width=\linewidth]{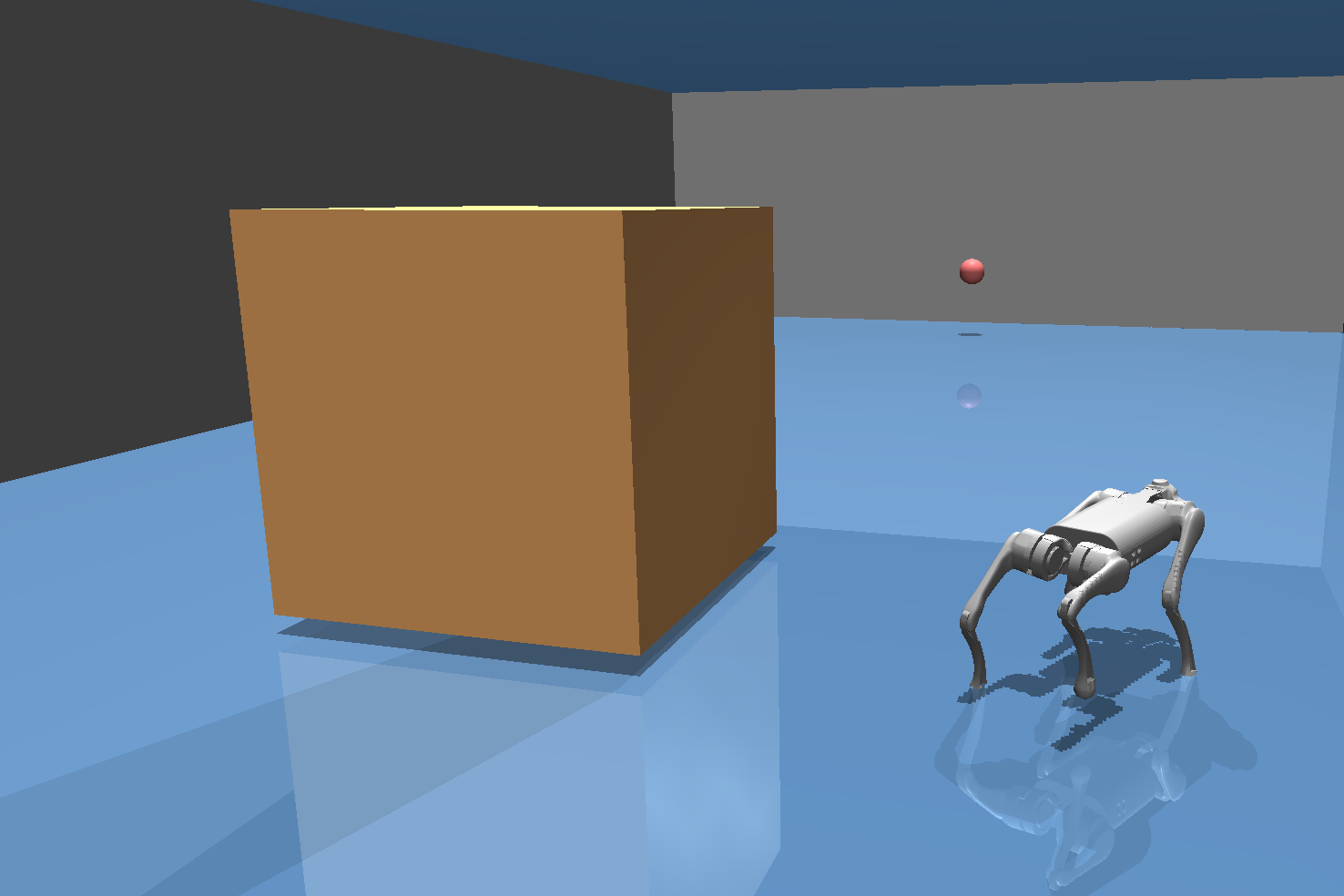}
        \caption{Avoiding the obstacle.}
        \label{fig:image2}
    \end{subfigure}
    \hfill
    \begin{subfigure}{0.32\textwidth}
        \centering
        \includegraphics[width=\linewidth]{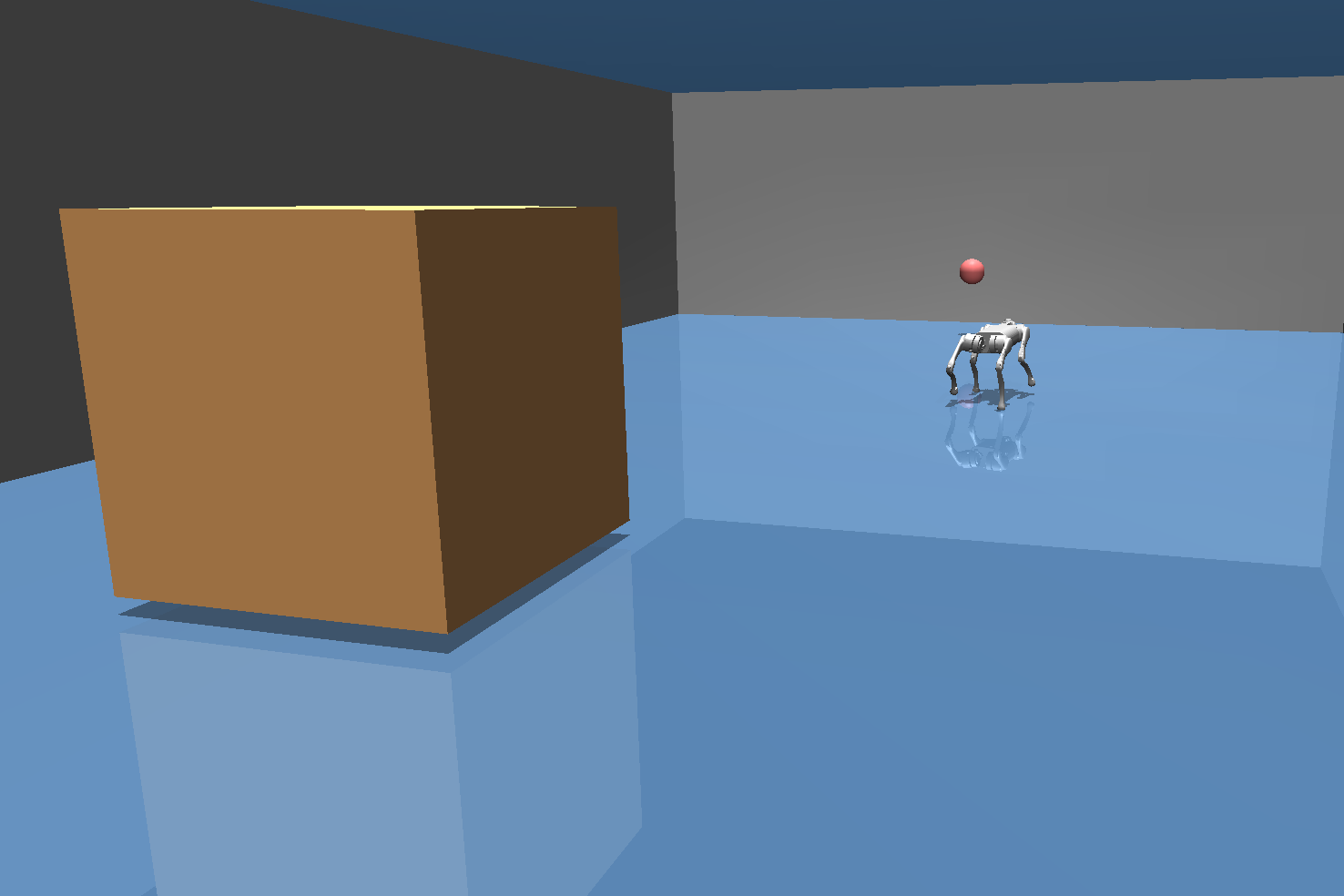}
        \caption{Reaching the target.}
        \label{fig:image3}
    \end{subfigure}
    \caption{Sequence of a robot Unitree GO1 using Hi-QARL to navigate a room with a moving obstacle (yellow box) towards a target position (red dot).}
    \label{fig:three_images}
\end{figure}
\section{Introduction}
    Effective locomotion in changing and unseen environment remains an ongoing challenge within the field of quadruped robots. In order to collaborate with humans, quadrupeds must safely navigate environments without causing damage to humans or other moving obstacles. The ultimate goal of robust robot locomotion is to find a safe path to target locations in the presence of changing environmental factors. State of the art approaches to quadruped locomotion have produced impressive results in terms of robustness to changing terrains and learning desired gaits~\cite{margolis2022rapidlocomotionreinforcementlearning, cheng2023extremeparkourleggedrobots, shafiee2023puppeteermarionettelearninganticipatory,  rudin2022learningwalkminutesusing, smith2023growlimitscontinuousimprovement}. However, modern approaches often fail when tested in out-of distribution settings with moving obstacles.
    
    One common approach to endow RL agents with robustness w.r.t. unexpected domain irregularities is to train them in the presence of adversarial perturbations from another agent~\cite{schott2024robust,pinto2017robust, cai2018curriculum} in a two-player zero-sum minimax game. The idea is that these perturbations will emulate the difficulties induced by mismatches between the simulated and real setting, thus giving the agent the experience required to perform well in a Sim2Real scenario. The solutions of this game are Nash equilibria, i.e., the saddle points of the 2-player min-max optimization problem. While often successful, the highly non-concave non-convex objective functions typically present in deep RL control problems, typically results in convergence to a suboptimal Nash equilibrium for the primary agent. 

    In this work, we propose a method to endow a quadruped with robustness to unseen dynamic obstacles in its navigation space. We adopt an adversarial RL perspective modelling the obstacle as a \textit{sentient} agent that act adversarially w.r.t. the quadruped, i.e., the protagonist agent. Crucially, we use a hierarchical controller for the protagonist agent, where the high-level navigation policy is trained to solve a curriculum over increasingly stronger adversaries. To model the strength of the adversary, we resort to the theory of quantal response equilibria (QRE), i.e., a generalisation of the Nash equilibrium that accounts for bounded rationality of the agents, thus relaxing the highly non-concave non-convex optimisation landscape and facilitating finding effective navigation policies~\cite{goeree2016quantal}. For this reason, we call our method Hierarchical policies via Quantal response Adversarial Reinforcement Learning (\textbf{Hi-QARL}).
 
\section{Related work}
Over the last few years, Reinforcement Learning (RL) has been demonstrated to be an effective framework for learning locomotion control policies. Most of these works focus on obtaining velocity-conditioned-locomotion policies that can be remotely controlled by a human operation using a joystick. However, these control policies lack the ability to navigate in cluttered, unstructured environments without human supervision~\cite{margolis2022rapidlocomotionreinforcementlearning,rudin2022learningwalkminutesusing,smith2022walkparklearningwalk,margolis2022walkwaystuningrobot,shafiee2023puppeteermarionettelearninganticipatory, sun2024learningbasedhierarchicalcontrolemulating}
However, recent work study the potential  deep RL to control locomotion and navigation simultaneously~\cite{rudin2022advancedskillslearninglocomotion,Lee_2024,cheng2023extremeparkourleggedrobots,kareer2023vinlvisualnavigationlocomotion,sombolestan2023hierarchicaladaptivelocomanipulationcontrol}.


Several works formulate a standard RL problem as a game between agents in order to maximise some notion of robustness. Notably, the robust adversarial RL setting (RARL)~\cite{pinto2017robust} formulates a two-player zero-sum Markov game between a protagonist and adversarial agent with opposing goals. By training the protagonist agent in a locomotion problem setting while facing adversarial perturbations applied by the adversary, the protagonist learns a policy that is more robust to domain transfer irregularities such as different frictions or masses.

\section{Preliminaries}

We consider two-player Markov games formulated as a Markov decision process (MDP) $\mathcal{M}=\langle\mathcal{S},\mathcal{O}_1,\mathcal{O}_2,\mathcal{A}_1,\mathcal{A}_2,\mathcal{R},\mathcal{P},\gamma,\iota\rangle$, where $\mathcal{S}$ is the state space, $\mathcal{O}_1$ and $\mathcal{O}_2$ are the continuous observation spaces of the two agents, $\mathcal{A}_1$ and $\mathcal{A}_2$ are the continuous action spaces of the agents,  $\mathcal{P}:\mathcal{S}\times\mathcal{A}_1\times\mathcal{A}_2\times\mathcal{S}\to\mathbb{R}$ is a transition probability density, $\mathcal{R}$ is a reward function, $\gamma\in[0,1)$ is a discount factor, and $\iota$ a probability density function over initial states. In a two-player entropy-regularised zero-sum Markov game, one agent seeks to maximise J, the other agent seeks to minimise J, and they both seek to maximise their policy entropy. The policies $\mu$ and $\nu$ of the agents aim to optimise the following entropy-regularised saddle-point optimisation

\begin{equation}
    J_{\mu^*,\nu^*}=\max_\mu \min_\nu J_{\mu,\nu} + \beta \mathcal{H}(\mu) - \alpha \mathcal{H}(\nu),\label{E:qarl_objective}
\end{equation}

where $J_{\mu,\nu}$ is the discounted return of the protagonist agent, $\mathcal{H}$ is the Shannon entropy of a distribution, and $\alpha,\beta \in \mathbb{R}^+$ are the temperature coefficients of the protagonist and adversary respectively. Our approach is based on solving the bounded rationality QRE modelled by the entropy-regularised saddle-point optimisation. Crucially, entropy-regularised RL~\cite{haarnoja2018softactorcriticoffpolicymaximum} provides a theoretical basis by which Markov games under bounded rationality can be solved. 

 \section{Method}
 We propose to train a navigation-locomotion hierarchical policy against an adversarial agent represented by a moving obstacle. By doing this, we aim to endow the agent with the ability to avoid static and dynamic obstacles in various configurations. The protagonist agent is a \textbf{Unitree GO1 quadruped} and the adversarial agent is a \textbf{moving obstacle}. Bot are controlled by a different soft-actor critic (SAC) agent~\cite{haarnoja2018softactorcriticoffpolicymaximum}. Unlike the protagonist agent whose rationality is maximal and fixed, we tune the rationality of the adversarial agent using its temperature $\alpha$. We start by solving the QRE with a highly random adversary, i.e., a high value for $\alpha$. By gradually decreasing the temperature $\alpha\to0$, we obtain a robust protagonist to the worst-case adversary as the maximal robustness target. In practice, we anneal the temperature of the adversary using the paradigm of self-paced deep RL~\cite{klink2020self, klink2021probabilistic, reddi2023robust} for automatic curricula generation. We model the temperature using a probability distribution $p_\omega(\alpha)$ defined by a distribution parameter vector $\omega$. Given $\alpha\in\mathbb{R}^+$, a suitable distribution for the adversarial temperature is the Gamma distribution $\Gamma(k,\theta)$, where $\omega=\langle k,\theta\rangle$ are the parameters of the distribution. We solve the following optimisation problem

\begin{equation}
    \min_\omega \, D_\text{KL}\left(p_\omega(\alpha) \, || \, \mu(\alpha)\right) 
    \quad \text{s.t.} \quad \mathbb{E}_{p_\omega(\alpha)}\left[J_{\mu,\nu}(\alpha)\right] \geq \xi, \quad D_\text{KL}\left(p_\omega(\alpha)||p_{\omega'}(\alpha)\right) \leq \epsilon.
    \label{E:spdl_qarl}
\end{equation}

where $\omega'$ is the parameter vector before the update and $\mu$ is a target temperature distribution. At each stage of the curriculum, the KL-divergence between the current distribution $p_\omega$ and the target $\mu$ is minimised. We select $\mu$ as a sharp Gamma distribution $\mu=\Gamma(1,10^{3})$, approximating a Dirac distribution centered on $\alpha=0$. The first constraint precludes catastrophic performance drops due to an overly strong adversary by only progressing the curriculum when the protagonist agent has reached a sufficient performance threshold $\xi$. The second constraint avoids abrupt changes in the distribution $p_\omega$ using a step-size constraint $\epsilon$. Note that the temperature $\beta$ of the protagonist is updated using the default SAC update~\cite{haarnoja2018softactorcriticoffpolicymaximum}.

\subsection{Hierachical policies}
We use a two-level hierarchical policy, as shown in Figure~\ref{hierachical_policy}, with two neural networks: the low-level policy $\pi^\text{loc}{\theta}(s^\text{loc}|v^{\text{cmd}})$ and the high-level policy $\pi^\text{nav}{\phi}(s^\text{nav})$. The low-level policy learns locomotion tasks, similar to prior work~\cite{rudin2022learningwalkminutesusing, margolis2023walk, margolis2022rapidlocomotionreinforcementlearning}, by training a robust velocity-conditioned controller using proximal policy optimization (PPO)~\cite{schulman2017proximalpolicyoptimizationalgorithms}, reward scheduling, and domain randomization to enhance agility and robustness. The scheduling adjusts penalty terms in the reward function, allowing the quadruped to learn stable movements and refine them over time. Domain randomization includes randomizing velocity commands $v^\text{cmd}$ and applying perturbations to the robot's center of mass velocities, helping the robot learn to anticipate sudden velocity changes for more stable gaits. The high-level policy handles navigation and obstacle avoidance, trained using SAC~\cite{haarnoja2018softactorcriticoffpolicymaximum}.\\
\textbf{State space:} The state space is divided into the navigation observation space and the locomotion navigation space. The navigation observation space has a history of the robot's position, orientation, angular and linear velocities, the relative distance to the target, and the relative distance to the closest distance to the nearest obstacle bounding box. The locomotion observation space includes the proprioception, the IMU information (velocimeter and gyroscope), the projected gravity on the torso center of mass, the torso height w.r,.t the floor and metrics related to the foot contact state.\\
\textbf{Action space:} Hi-QARL uses three action spaces:
\begin{itemize}
  \item \textbf{Navigation Action:} Desired command velocities: $a^\text{nav}_t \rightarrow \{v^\text{cmd}_{x,t}, v^\text{cmd}_{y,t}, \omega^\text{cmd}_{z,t}\} \in [-1,1]^3$
  
  \item \textbf{Locomotion Action:} Desired joint positions: $a^\text{loc}_t  \rightarrow q^\text{des}_t \in \mathbb{R}^{12}$
  
  \item \textbf{Adversary Action:} Desired obstacle lateral velocity: $a^\text{adversary}_t \rightarrow v^\text{obstacle}_{y,t} \in [-1,1]$
\end{itemize}

\textbf{Reward functions:} Each of the policies from Hi-QARL follows a different reward function. The low-level policy's reward function consists of two types of rewards: task rewards (for tracking angular and linear velocities) and auxiliary penalties for stability, including pitch, roll, and vertical velocity penalties, self-collision and joint limit penalties, and penalties for smoothness, gait style, and posture. The navigation policy reward drives the robot toward the target, corrects heading, penalizes stalling far from the destination, and discourages collisions. A bonus term rewards staying at the target. Finally, the adversary reward seeks to minimize the distance between the robot and obstacles. The reward functions are described in detail in Appendix~\ref{appendix:rewards}.


\section{Experimental results}
We benchmark our trained agents against a multi-obstacle navigation problem containing a pre-trained highly rational adversarial obstacle and two randomly placed static obstacles. Note that the layout of the maze in the benchmark environment is different to that of the training environment, since it contains additional two static obstacles that appear in random positions each episode.

\begin{figure}[h]
    \centering
    \begin{minipage}{0.34\textwidth}
        \centering
        \includegraphics[width=\textwidth]{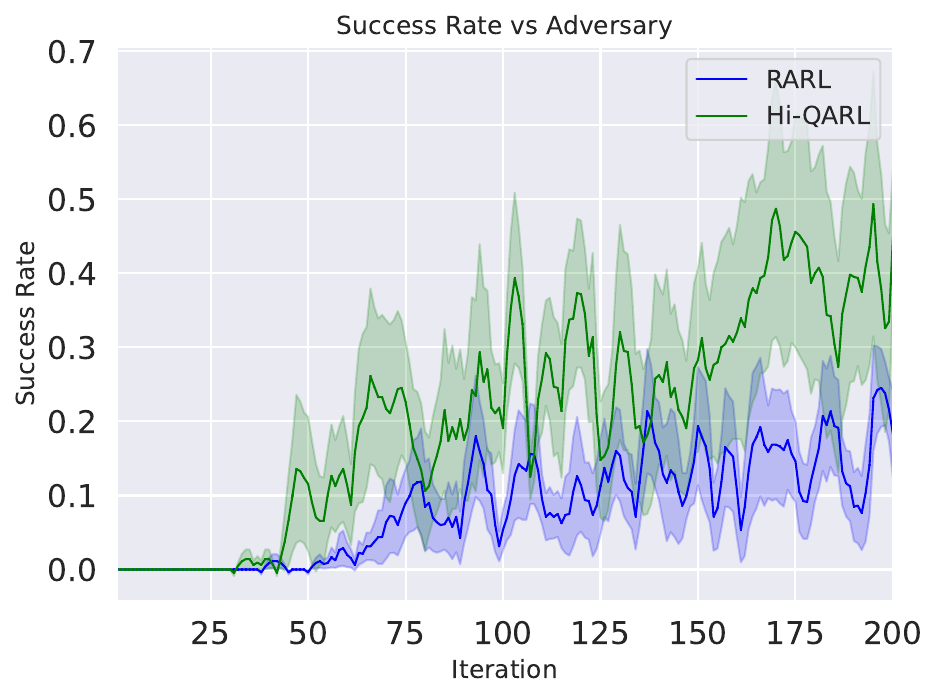}
        \caption{Online success rate of RARL vs Hi-QARL. }
        \label{F:exp_perf}
    \end{minipage}
    \hfill
    \begin{minipage}{0.32\textwidth}
        \centering
        \includegraphics[width=\textwidth]{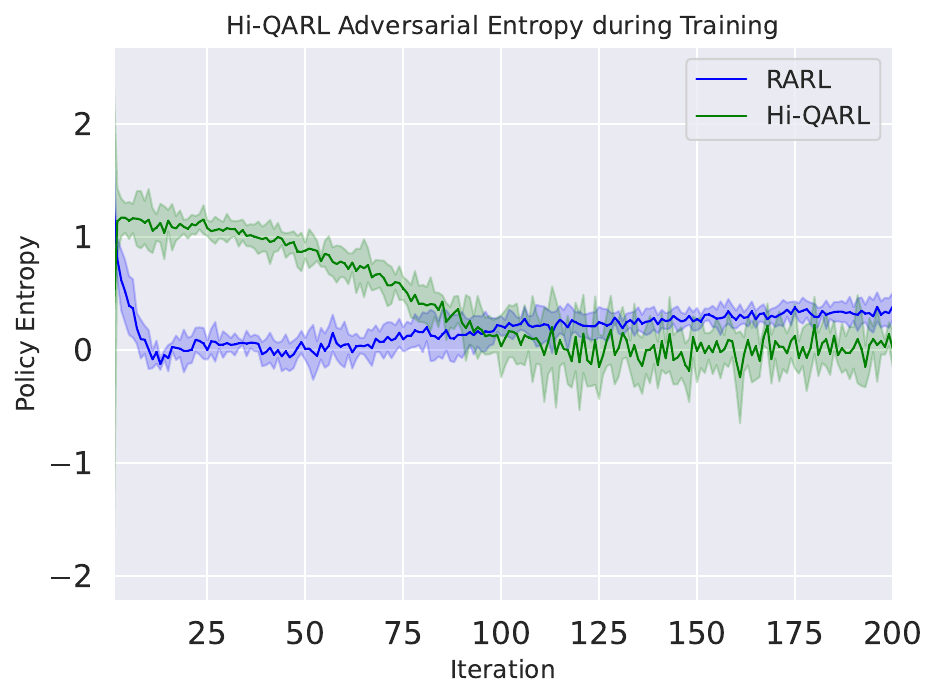}
        \caption{Adversarial policy entropy of RARL and Hi-QARL during training.}
        \label{F:exp_ent}
    \end{minipage}
    \hfill
    \begin{minipage}{0.32\textwidth}
        \centering
        \includegraphics[width=\textwidth]{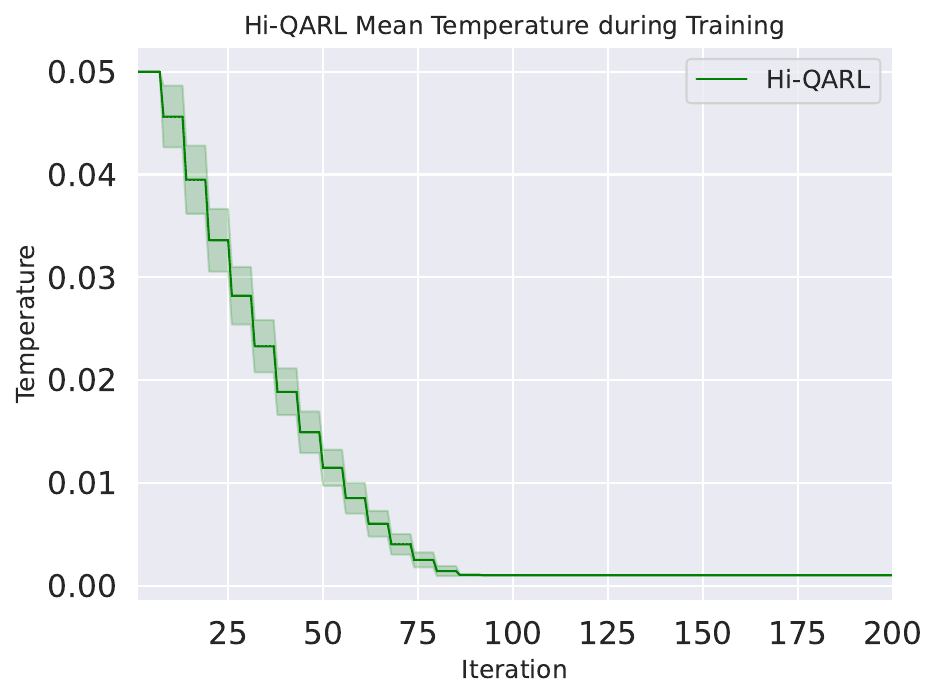}
        \caption{Mean temperature of Hi-QARL adversary during training.}
        \label{F:exp_temp}
    \end{minipage}
\end{figure}

We aim to show that training against the adversary endows Hi-QARL with robustness to these unseen environmental obstacles than the other methods. In addition to Hi-QARL, we benchmark RARL (the same adversarial setting but without entropy regularisation), Static (a quadruped agent trained against a randomly placed static obstacle), and Random (a quadruped agent trained against an obstacle with random movements). Figure~\ref{T:exp_results} shows the performance for the multi-obstacle problem.
\begin{wraptable}{r}{4.5cm}
    \tabcolsep=0.1cm
    \resizebox{4.5cm}{!}{
    \begin{tabular}{c|cc}
        \hline
        Algorithm & Robustness\\
        \hline
        Static  & $0.0\pm0.0\%$\\
        Random & $0.0\pm0.0\%$\\
        RARL & $0.13\pm0.01\%$\\
        \hline \hline
        Hi-QARL (ours)  & $\mathbf{0.39\pm0.09\%}$\\
        \hline
    \end{tabular}
    }
    \caption{Robustness to random obstacles across $10$ seeds}\label{T:exp_results}
\end{wraptable}
Figure~\ref{F:exp_perf} shows the success rate of RARL and Hi-QARL against their respective adversaries as training progresses (the Static and Random benchmarks are not compared here as their obstacles are not controlled by learning agents). Hi-QARL can clearly be seen to outperform RARL during training due to the ease of the complexity of the adversarial RL problem. Figure~\ref{F:exp_ent} compares the entropy of the adversarial agent, where the gradual decrease in Hi-QARL clearly indicates the controlled increase in rationality. By contrast, the adversary in RARL updates its temperature using the typical SAC update~\cite{haarnoja2018softactorcriticoffpolicymaximum}. Figure~\ref{F:exp_temp} demonstrates the average temperature sampled from $p_\omega(\alpha)$ as training progresses. Note that the curriculum progresses subject to the performance of the protagonist, which increases sooner than the success rate in Figure~\ref{F:exp_perf}. For the sake of reproducibility, Appendix~\ref{appendix:experiments} describes our experimental setting in detail.

\section{Conclusion}
\label{sec:conclusion}

We presented Hi-QARL, an algorithm leveraging bounded rationality curriculum adversarial learning and a hierarchical reinforcement learning (RL) controller to learn a quadruped locomotion policy robust to moving obstacles. By combining the benefits of adversarial learning with the ease of solving a maximum entropy RL problem, we show a reliable increase in testing performance of our method over existing baselines. Despite our positive early findings, our method can only manage one obstacle at a time in front of the robot. In future work, we plan to extend our method so our robot can safely navigate across an arbitrary number of moving obstacles. 


\clearpage


\section*{Acknowledgments}
This work was funded by the German Federal Ministry of Education and Research (BMBF) (Project:
01IS22078). This work was also funded by Hessian.ai through the project ’The Third Wave of
Artificial Intelligence – 3AI’ by the Ministry for Science and Arts of the state of Hessen.  Calculations for this research were conducted on the Lichtenberg high-performance computer of the TU Darmstadt and the Intelligent Autonomous Systems (IAS) cluster at TU Darmstadt and with the HPC resources provided by the Erlangen National High Performance Computing Center (NHR@FAU) of the Friedrich-Alexander-Universitat Erlangen-N\"{u}rnberg (FAU) under the NHR project b187cb. NHR funding is provided by federal and Bavarian state authorities. NHR@FAU hardware is partially funded by the German Research Foundation (DFG) – 440719683.

\bibliography{corl}  

\newpage

\appendix

\section{Additional results}
\label{appendix:results}

\begin{figure}[h!]
    \centering
    \begin{subfigure}{0.22\textwidth}
        \centering
        \includegraphics[width=\linewidth]{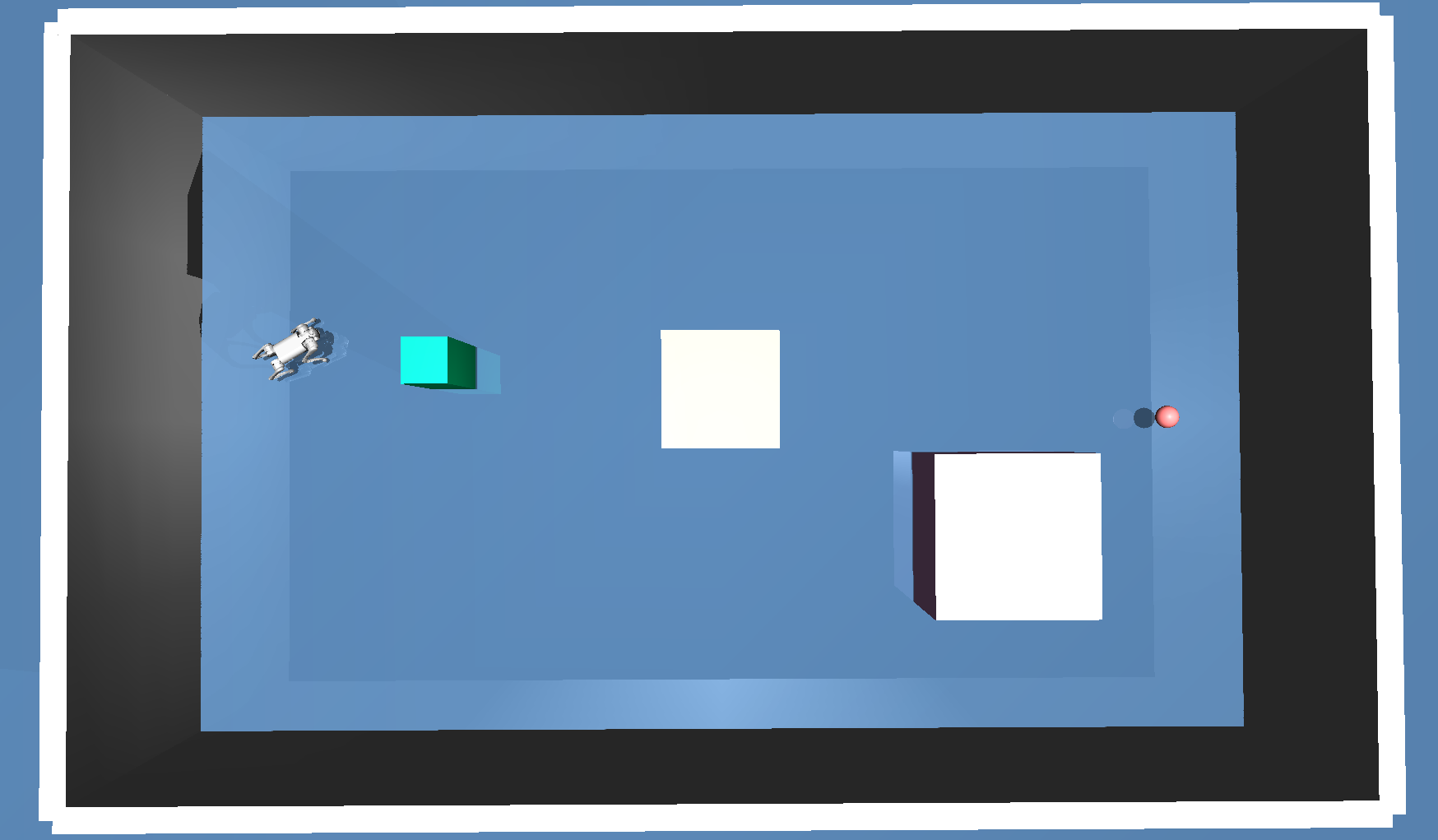}
        \label{fig:image1}
    \end{subfigure}
    \begin{subfigure}{0.22\textwidth}
        \centering
        \includegraphics[width=\linewidth]{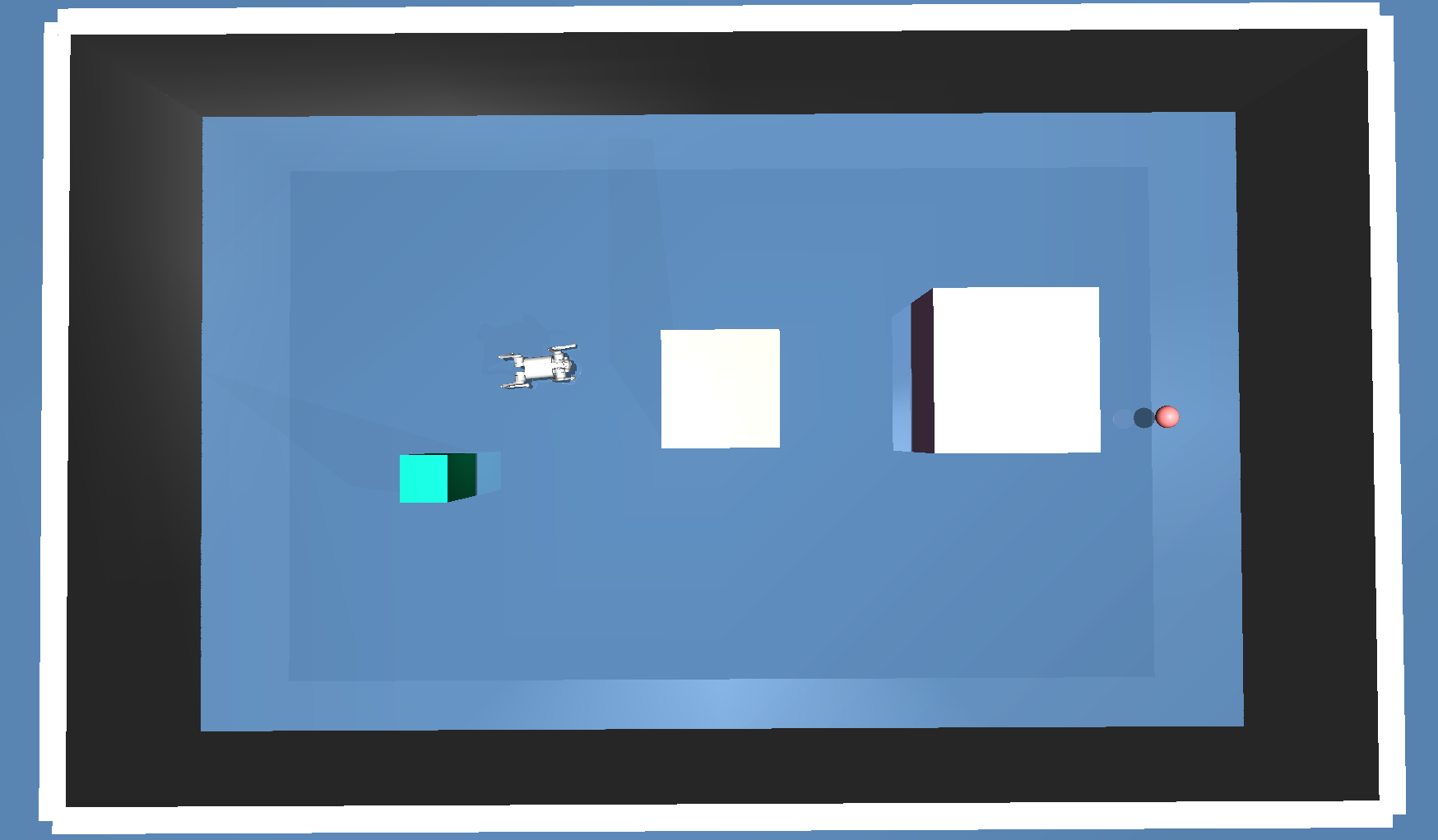}
        \label{fig:image2}
    \end{subfigure}
    \begin{subfigure}{0.22\textwidth}
        \centering
        \includegraphics[width=\linewidth]{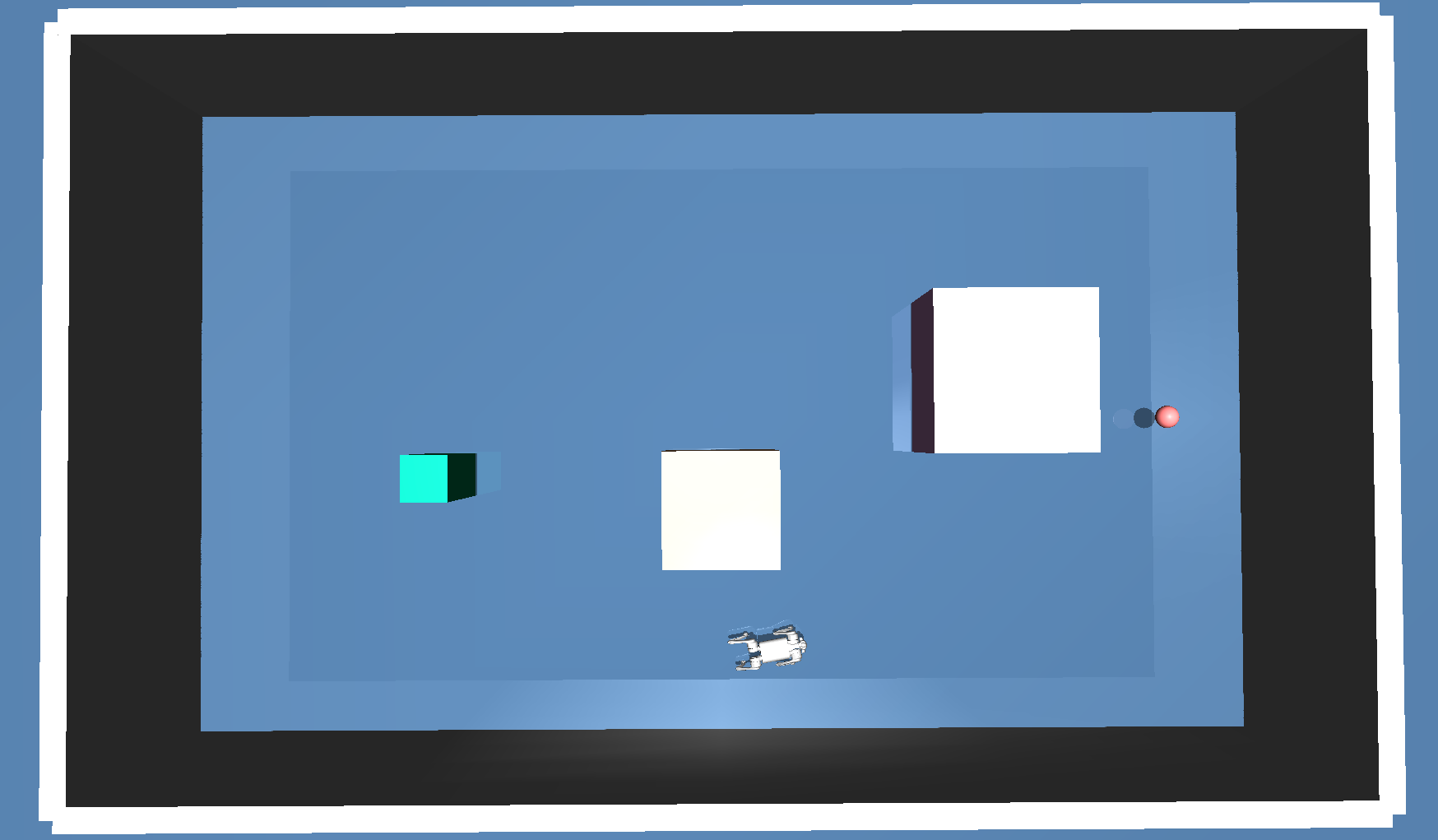}
        \label{fig:image3}
    \end{subfigure}
    \begin{subfigure}{0.22\textwidth}
        \centering
        \includegraphics[width=\linewidth]{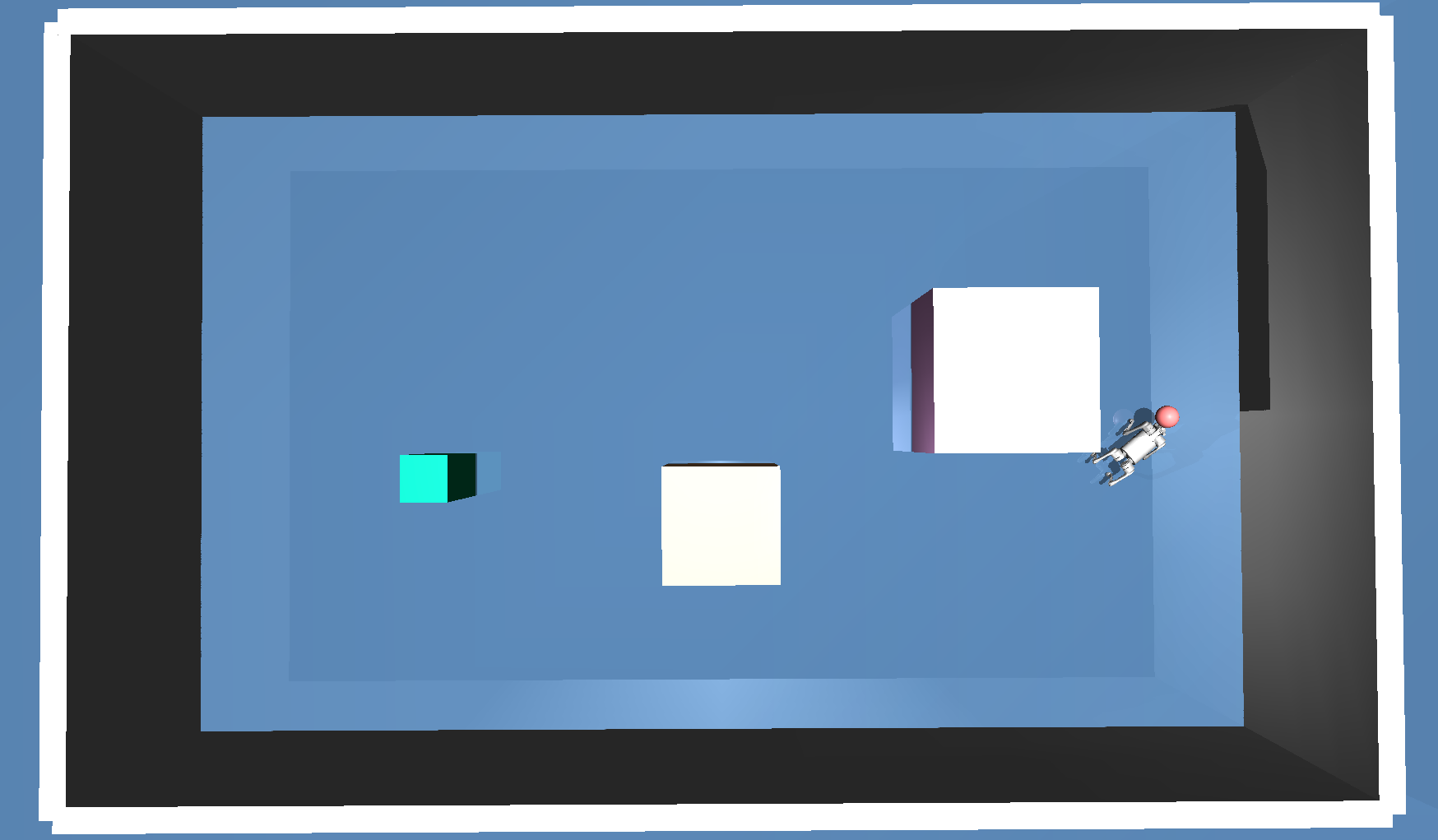}
        \label{fig:image4}
    \end{subfigure}
    \caption{GO1 robot successfully navigates a new maze with two new unseen static obstacles in addition to a moving obstacle.}
    \label{final_results}
\end{figure}

\section{Experiments details}
\label{appendix:experiments}

The methods were trained with an NVIDIA GeForce RTX 4090 GPU. The simulation environment is MuJoCo~\cite{todorov2012mujoco}, and we use the Unitree GO1 model provided by MuJoco Menagerie~\cite{menagerie2022github}. In addition, all our experiments use MushroomRL~\cite{JMLR:v22:18-056} as the main framework. Lastly, our models are implemented in PyTorch.

\subsection{Extension hierarchical policies}
\begin{figure}[h!]
    \centering
    \includegraphics[width=.6\textwidth]{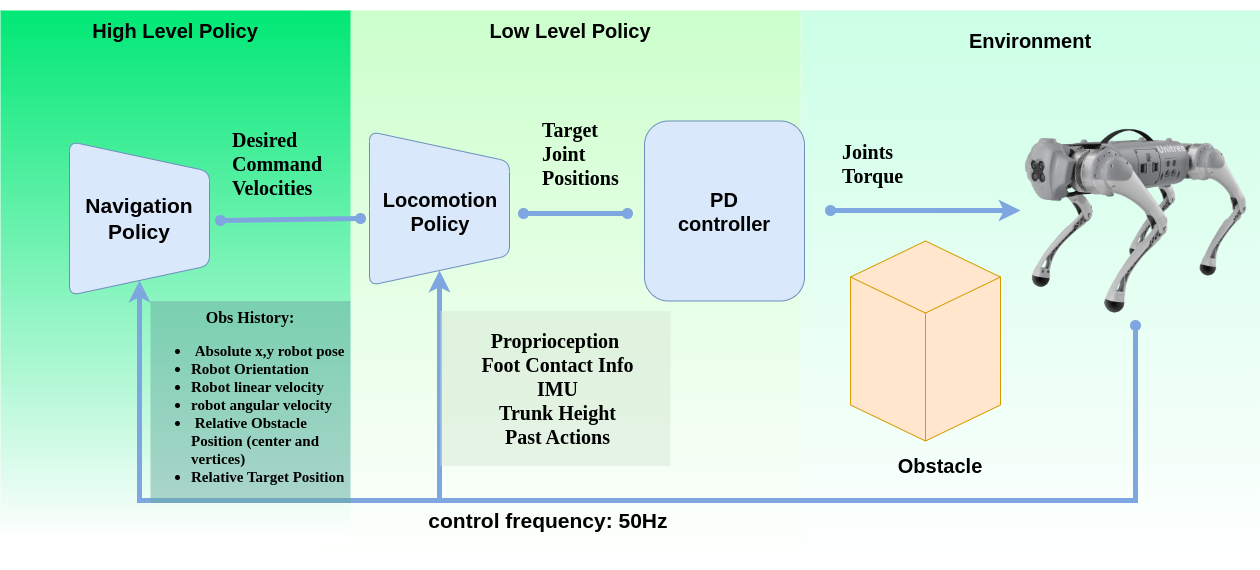}
    \caption{The hierarchical policy has a navigation policy and a locomotion policy.}
    \label{hierachical_policy}
\end{figure}

\subsection{Training parameters and rewards}
\begin{table}[h]
\centering
\begin{tabular}{|>{\centering\arraybackslash}m{10.5cm}|}
\hline
\textbf{PPO} \\
\hline
\end{tabular}

\begin{tabular}{|>{\centering\arraybackslash}m{5cm}|>{\centering\arraybackslash}m{5cm}|}
\hline
\textbf{Parameters} & \textbf{Values} \\
\hline
horizon & 1000\\
\hline
$\gamma$ & 0.99 \\
\hline
mini batch & 64 \\
\hline
batch & 2048 \\
\hline
lam & 0.95 \\
\hline
ratio clipping & 0.2 \\
\hline
target kl & 0.03 \\
\hline
entropy coefficient & 0. \\
\hline

\end{tabular}

\vspace{0.5cm}
\caption{Locomotion policy hyperparameters.}
\label{table:paramsPPO}
\end{table}

\begin{table}[h]
\centering
\begin{tabular}{|>{\centering\arraybackslash}m{10.5cm}|}
\hline
\textbf{SAC} \\
\hline
\end{tabular}

\begin{tabular}{|>{\centering\arraybackslash}m{5cm}|>{\centering\arraybackslash}m{5cm}|}
\hline
\textbf{Parameters} & \textbf{Values} \\

\hline
horizon & 1000\\
\hline
$\gamma$ & 0.99 \\
\hline
warmup transitions & 3000 \\
\hline
batch & 256 \\
\hline
max replay buffer & 1e6 \\
\hline
$\tau$ & 0.005\\
\hline
learned entropy & True\\
\hline
learning rate entropy & 3e-4 \\
\hline
learning rate policy/critic & 3e-4 \\
\hline
min and max value for policy log std & -20, 2 \\
\hline
\end{tabular}

\vspace{0.5cm}
\caption{Navigation and adversarial policy hyperparameters.}
\label{table:paramsSAC}
\end{table}
\begin{table}[h!]
\centering
\begin{tabular}{|>{\centering\arraybackslash}m{5cm}|>{\centering\arraybackslash}m{5cm}|>{\centering\arraybackslash}m{2cm}|}
\hline
\textbf{Reward Name} & \textbf{Reward Expression} & \textbf{Weights} \\
\hline
Velocity tracking & $e^{-\frac{|v_{xy} - v_{xy}^{cmd}|}{\sigma_{xy}} }$ & 2 \\
\hline
Angular velocity tracking & $e^{-\frac{| \omega_z - \omega_{z}^{cmd}|}{\sigma_z}}$ & 1 \\
\hline
Height penalty & $\| h_z - h_z^{*} \|$ & -30 \\
\hline
Self-collision Penalty & $\mathbbm{1}_{\text{selfcollision}}$ & -0.7 \\
\hline
Joint-limit-violation penalty & $\mathbbm{1}_{q_i > q_{max} \lor q_i < q_{min}}$ & -10 \\
\hline
Base Orientation & $\left| g_{xy} \right|^2$ & 0.4 \\
\hline
Roll-Pitch velocity Penalty & $\omega^2_{xy}$ & -1 \\
\hline
Z velocity penalty & $v^2_z$ & 2 \\
\hline
Joint torques & $|\tau|^2$ & -2e-4 \\
\hline
Joint accelerations & $|\Ddot{q}|^2$ & -2.5e-7 \\
\hline
Action rate & $|a_{t-1}-a_{t}|^2$ & -0.02 \\
\hline
Feet airtime & $\sum{t_{air}} * \mathbbm{1}_{\text{new contact}}$ & -0.1 \\
\hline
Feet symmetry contact & \[
f_{\text{symmetry}} = \left| c_{\text{FL}} - c_{\text{RR}} \right| + \left| c_{\text{FR}} - c_{\text{RL}} \right|
\] & -0.5 \\
\hline
\end{tabular}
\vspace{0.5cm}

\caption{Rewards for locomotion policy with expressions and weights, which are scaled by $0.02$.}
\label{table:reward_loco}
\end{table}
\vspace{-.5cm}
\begin{table}[h!]
\centering
\begin{tabular}{|>{\centering\arraybackslash}m{5cm}|>{\centering\arraybackslash}m{5cm}|>{\centering\arraybackslash}m{2cm}|}
\hline
\textbf{Reward Name} & \textbf{Reward Expression} & \textbf{Weights} \\
\hline
Moving to target reward & $\vec{heading_{robot}} \cdot \vec{velocity_{robot}}$ & 1 \\
\hline
Heading reward & $e^{| \text{yaw} - \text{yaw}^{target}|}$ & 0.5 \\
\hline
Reach target bonus &$1 \quad \text{if } d_{\text{robot,target}} <= 0.5 $ & 10 \\
\hline
Collision penalty &  $-10 * 1_{obstacle\_collision}  -5 * 1_{walls\_collision}$ & 1 \\
\hline
Stall penalty & $1 \quad \text{if } d_{\text{robot,target}} > 0.5 \text{ and } v_{xy} < 0.1$

 & -10 \\
\hline
\end{tabular}
\vspace{0.5cm}
\caption{Rewards table for the navigation policy with expressions and weights.}
\label{table:reward_nav}
\end{table}
\vspace{-.5cm}
\begin{table}[h!]
\centering
\begin{tabular}{|>{\centering\arraybackslash}m{5cm}|>{\centering\arraybackslash}m{5cm}|>{\centering\arraybackslash}m{2cm}|}
\hline
\textbf{Reward name} & \textbf{Reward expression} & \textbf{Weights} \\
\hline
Antagonist position tracking &  $e^{-| \text{pos}^\text{y,robot} - \text{pos}^{\text{y,obstacle}}|}$ & 1 \\
\hline
\end{tabular}
\vspace{0.5cm}
\caption{Rewards table for the antagonist policy with expressions and weights.}
\label{table:reward_adv}
\end{table}
\end{document}